\documentclass[acmtog,nonacm]{acmart} 

\makeatletter
\let\@authorsaddresses\@empty
\makeatother

\renewcommand\footnotetextcopyrightpermission[1]{} 
\usepackage{colortbl}
\usepackage{graphicx} 
\usepackage[ruled]{algorithm2e} 

\SetAlFnt{\small}
\SetAlCapFnt{\small}
\SetAlCapNameFnt{\small}
\SetAlCapHSkip{0pt}
\usepackage{multirow}
\usepackage{booktabs} 
\usepackage{color}
\definecolor{applegreen}{rgb}{0.55, 0.71, 0.0}
\definecolor{autumnorange}{rgb}{0.87, 0.61, 0.33}
\definecolor{moreprompt}{HTML}{C2D9FF}
\definecolor{oneprompt}{HTML}{FFE4D6}

\newcommand{\ourgenerator}{3DPortraitGAN\resizebox{0.30cm}{!}{~\protect
\begingroup%
  \makeatletter%
  \providecommand\color[2][]{%
    \errmessage{(Inkscape) Color is used for the text in Inkscape, but the package 'color.sty' is not loaded}%
    \renewcommand\color[2][]{}%
  }%
  \providecommand\transparent[1]{%
    \errmessage{(Inkscape) Transparency is used (non-zero) for the text in Inkscape, but the package 'transparent.sty' is not loaded}%
    \renewcommand\transparent[1]{}%
  }%
  \providecommand\rotatebox[2]{#2}%
  \newcommand*\fsize{\dimexpr\f@size pt\relax}%
  \newcommand*\lineheight[1]{\fontsize{\fsize}{#1\fsize}\selectfont}%
  \ifx\svgwidth\undefined%
    \setlength{\unitlength}{83.4210115bp}%
    \ifx\svgscale\undefined%
      \relax%
    \else%
      \setlength{\unitlength}{\unitlength * \real{\svgscale}}%
    \fi%
  \else%
    \setlength{\unitlength}{\svgwidth}%
  \fi%
  \global\let\svgwidth\undefined%
  \global\let\svgscale\undefined%
  \makeatother%
  \begin{picture}(1,0.74034917)%
    \lineheight{1}%
    \setlength\tabcolsep{0pt}%
    \put(0,0){\includegraphics[width=\unitlength,page=1]{pyramid.pdf}}%
  \end{picture}%
\endgroup%
}\xspace}

\newcommand{\ourname}{Portrait3D\xspace}

\acmJournal{TOG}
\acmYear{2024}
\acmMonth{7}

\begin{document}
\title{\ourname: Text-Guided High-Quality 3D Portrait Generation Using Pyramid Representation and GANs Prior}


\author{Yiqian Wu}
\affiliation{%
     \institution{State Key Lab of CAD\&CG, Zhejiang University}
     \city{Hangzhou}
     \country{China}
     }
\orcid{0000-0002-2432-809X}
\email{onethousand@zju.edu.cn}

\author{Hao Xu}
\affiliation{%
     \institution{State Key Lab of CAD\&CG, Zhejiang University}
     \city{Hangzhou}
     \country{China}
     }
\orcid{0000-0001-5690-367X}
\email{haoxu38@outlook.com}

\author{Xiangjun Tang}
\affiliation{%
     \institution{State Key Lab of CAD\&CG, Zhejiang University}
     \city{Hangzhou}
     \country{China}
     }
\orcid{0000-0001-7441-0086}
\email{xiangjun.tang@outlook.com}

\author{Xien Chen}
\affiliation{%
     \institution{Yale University}
     \city{New Haven}
     \country{United States of America}
     }
\orcid{0009-0000-7397-4955}
\email{xien.chen@yale.edu}

\author{Siyu Tang}
\affiliation{%
     \institution{ETH Zürich}
     \city{Zürich}
     \country{Switzerland}
     }
\orcid{0000-0002-1015-4770}
\email{siyu.tang@inf.ethz.ch}

\author{Zhebin Zhang}
\affiliation{%
     \institution{OPPO US Research Center}
     \city{Bellevue}
     \country{United States of America}
     }
\orcid{0009-0008-2860-5112}
\email{zhebin.zhang@oppo.com}

\author{Chen Li}
\affiliation{%
     \institution{OPPO US Research Center}
     \city{Bellevue}
     \country{United States of America}
     }
\orcid{0009-0002-6140-9216}
\email{chen.li@oppo.com}

\author{Xiaogang Jin}
\authornote{Corresponding author.}
\affiliation{%
     \institution{State Key Lab of CAD\&CG, Zhejiang University}
     \city{Hangzhou}
     \country{China}
     }
\orcid{0000-0001-7339-2920}
\email{jin@cad.zju.edu.cn}

\begin{abstract}


Existing neural rendering-based text-to-3D-portrait generation methods typically make use of human geometry prior and diffusion models to obtain guidance. However, relying solely on geometry information introduces issues such as the Janus problem, over-saturation, and over-smoothing. We present \textit{\ourname}, a novel neural rendering-based framework with a novel joint geometry-appearance prior to achieve text-to-3D-portrait generation that overcomes the aforementioned issues.
To accomplish this, we train a 3D portrait generator, 3DPortraitGAN\resizebox{0.30cm}{!}{~\protect
\begingroup%
  \makeatletter%
  \providecommand\color[2][]{%
    \errmessage{(Inkscape) Color is used for the text in Inkscape, but the package 'color.sty' is not loaded}%
    \renewcommand\color[2][]{}%
  }%
  \providecommand\transparent[1]{%
    \errmessage{(Inkscape) Transparency is used (non-zero) for the text in Inkscape, but the package 'transparent.sty' is not loaded}%
    \renewcommand\transparent[1]{}%
  }%
  \providecommand\rotatebox[2]{#2}%
  \newcommand*\fsize{\dimexpr\f@size pt\relax}%
  \newcommand*\lineheight[1]{\fontsize{\fsize}{#1\fsize}\selectfont}%
  \ifx\svgwidth\undefined%
    \setlength{\unitlength}{83.4210115bp}%
    \ifx\svgscale\undefined%
      \relax%
    \else%
      \setlength{\unitlength}{\unitlength * \real{\svgscale}}%
    \fi%
  \else%
    \setlength{\unitlength}{\svgwidth}%
  \fi%
  \global\let\svgwidth\undefined%
  \global\let\svgscale\undefined%
  \makeatother%
  \begin{picture}(1,0.74034917)%
    \lineheight{1}%
    \setlength\tabcolsep{0pt}%
    \put(0,0){\includegraphics[width=\unitlength,page=1]{pyramid.pdf}}%
  \end{picture}%
\endgroup%
}\xspace, as a robust prior. This generator is capable of producing $360^{\circ}$ canonical 3D portraits, serving as a starting point for the subsequent diffusion-based generation process. 
To mitigate the ``grid-like'' artifact caused by the high-frequency information in the feature-map-based 3D representation commonly used by most 3D-aware GANs, we integrate a novel \textit{pyramid tri-grid} 3D representation into 3DPortraitGAN\resizebox{0.30cm}{!}{~\protect}\xspace.
To generate 3D portraits from text, we first project a randomly generated image aligned with the given prompt into the pre-trained 3DPortraitGAN\resizebox{0.30cm}{!}{~\protect}\xspace's latent space. The resulting latent code is then used to synthesize a \textit{pyramid tri-grid}. 
Beginning with the obtained \textit{pyramid tri-grid}, we use score distillation sampling to distill the diffusion model's knowledge into the \textit{pyramid tri-grid}. 
Following that, we utilize the diffusion model to refine the rendered images of the 3D portrait and then use these refined images as training data to further optimize the \textit{pyramid tri-grid}, effectively eliminating issues with unrealistic color and unnatural artifacts.
Our experimental results show that \ourname can produce realistic, high-quality, and canonical 3D portraits that align with the prompt.

\end{abstract}

%
%

%
%


\begin{teaserfigure}
  \centering
  \includegraphics[width=\linewidth]{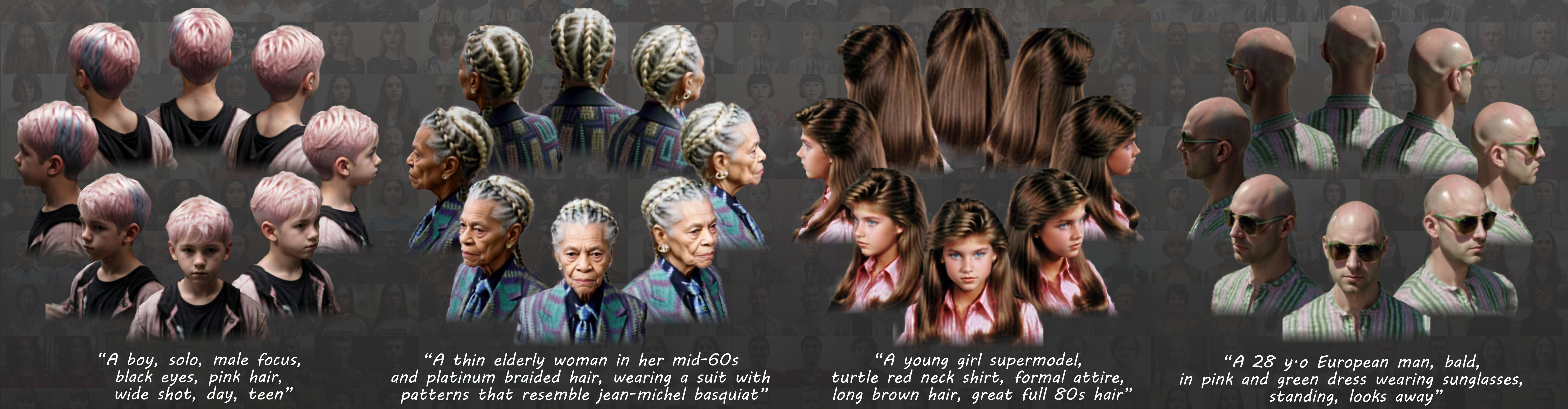}
  \caption{
  Using text as input, our text-to-3D-portrait method, \ourname, can automatically generate a variety of realistic textured 3D portraits. \ourname consistently produces high-quality 3D portraits that are aligned with the provided text prompts. 
  Each portrait is rendered from eight different views using volume rendering.
  }
\label{fig:teaser}
\end{teaserfigure}

\maketitle

\section{Introduction}
\label{sec: Introduction}

3D portrait modeling can benefit from neural rendering techniques, such as NeRF \cite{DBLP:conf/eccv/MildenhallSTBRN20}, NeuS \cite{DBLP:conf/nips/WangLLTKW21}, and Gaussian Splatting \cite{10.1145/3592433}, due to their powerful ability to represent detailed shapes and produce realistic rendering results.
The introduction of the diffusion model \cite{ho2020denoising} and score distillation sampling (SDS) \cite{DBLP:conf/iclr/PooleJBM23} has further spurred the rapid development of text-to-3D-portrait generation by distilling the knowledge of 2D image prior into 3D representation.
To obtain shape guidance, diffusion-based 3D portrait generation can leverage human geometry prior information, including SMPL \cite{DBLP:journals/tog/LoperM0PB15} (and its variants), FLAME \cite{DBLP:journals/tog/LiBBL017}, DensePose \cite{DBLP:conf/cvpr/GulerNK18}, and imGHUM \cite{Alldieck_2021_ICCV}.


Existing methods, however, primarily use geometric priors as initial points for generation in order to establish shape constraints, with the diffusion model providing all appearance information. This can result in issues like inconsistent or unrealistic textures. Furthermore, applying SDS to a coarse initialization can lead to over-saturation and over-smoothing. As a result, a robust joint geometry-appearance prior is still required to achieve realistic and high-quality 3D portrait generation. The key challenge in accomplishing this is the requirement for the prior to include full-head geometry-appearance information while also having the expressive power to portray a diverse range of 3D portraits.

In this paper, we present \textit{\ourname}, a neural rendering-based framework to generate high-quality and realistic 3D portraits following the input prompt. \ourname utilizes novel joint geometry-appearance prior information from 3D-aware GANs and guidance from the diffusion model.
However, performing score distillation sampling directly on the feature-map-based 3D representation used by many 3D-aware GANs can result in the ``grid-like'' artifact, which is caused by underlying exclusive high-frequency information.
To address this, we propose 3DPortraitGAN\resizebox{0.30cm}{!}{~\protect}\xspace, a 3D portrait generator utilizing a novel \textit{pyramid tri-grid} 3D representation to effectively alleviate the ``grid-like'' artifact. The symbol ``\resizebox{0.30cm}{!}{~\protect} (pyramid)'' indicates the usage of a pyramid-structure representation. 3DPortraitGAN\resizebox{0.30cm}{!}{~\protect}\xspace is capable of producing $360^{\circ}$ canonical 3D full-head portraits and providing a prior for our framework.
During the process of text-to-3D-portraits generation, given a text prompt, we first randomly generate a portrait image by feeding the text prompt into the diffusion model, and then project the generated image into the latent space of our generator. The resulting latent code is used to synthesize the corresponding \textit{pyramid tri-grid}, which serves as the starting point of the subsequent diffusion-based generation process. 
Following that, we distill the knowledge of the diffusion model into the \textit{pyramid tri-grid} through score distillation sampling. This process produces a 3D portrait that aligns with the input prompt.
To further enhance the quality of the obtained 3D portrait, we apply the diffusion model to process its rendered images. The refined rendered images are then used as training data to optimize the \textit{pyramid tri-grid}, yielding the final results.

Through comprehensive experiments, we demonstrate that \ourname is capable of generating high-quality, view-consistent, realistic, and canonical 3D portraits that align well with the input text prompt. 
Notably, our generation pipeline requires only a graphics card with 12GB of memory.

In summary, our work makes the following major contributions:
\begin{itemize}



    \item  A 3D portrait generator that employs a novel \textit{pyramid tri-grid} 3D representation to mitigate the ``grid-like'' artifact, and learns a distribution of $360^{\circ}$ canonical 3D portraits to serve as a robust prior for our framework.

    \item A text-to-3D-portrait generation framework, \ourname, leverages prior information from our 3D portrait generator and guidance from the diffusion model, producing high-quality, view-consistent, realistic, and canonical 3D portraits that are in alignment with the input text prompts.

\end{itemize}

\section{Related Work}
\label{sec: Related Work}

\subsection{3D-aware GANs}
Ever since Goodfellow et al.'s groundbreaking innovation of Generative Adversarial Networks (GANs) \cite{DBLP:conf/nips/GoodfellowPMXWOCB14} in 2014, a multitude of GAN models \cite{DBLP:journals/corr/RadfordMC15,DBLP:conf/nips/GulrajaniAADC17,DBLP:conf/iclr/KarrasALL18} have been proposed, yielding impressive outcomes in the domain of photorealistic image synthesis. 
Over recent years, 2D generators have been extended by incorporating voxel rendering \cite{DBLP:conf/nips/Nguyen-PhuocRMY20,DBLP:conf/nips/ZhuZZ00TF18} or NeRF rendering \cite{DBLP:conf/iclr/GuL0T22,DBLP:conf/cvpr/ChanMK0W21} with the generative network backbone, achieving view-consistent image generation and catalyzing the emergence of 3D-aware GANs.
A number of effective optimization strategies and frameworks \cite{Chen_2023_ICCV,DBLP:journals/corr/abs-2206-10535,DBLP:journals/corr/abs-2401-02411} have been developed  to facilitate the generation of high-resolution outcomes, eliminating the necessity for a super-resolution module.
The endeavor to reconstruct 3D faces from input images using 3D-aware generators has spurred the development of inversion techniques \cite{DBLP:conf/cvpr/XieOPLC23,DBLP:conf/cvpr/Yin0W0LGFCSOY23,DBLP:conf/wacv/KoCCRK23,DBLP:journals/tog/TrevithickCSCLYKCRN23}. This progression has enhanced the manipulation of real images and videos.

The 3D representations within 3D-aware GANs can be parameterized using coordinate-based networks \cite{DBLP:conf/cvpr/ChanMK0W21,DBLP:conf/iclr/GuL0T22}, feature maps \cite{DBLP:conf/cvpr/ChanLCNPMGGTKKW22,An_2023_CVPR}, signed distance fields (SDF) \cite{NEURIPS2022_cebbd24f}, radiance manifolds \cite{DBLP:conf/siggrapha/0012XXWCY023},
or voxel grids \cite{DBLP:conf/nips/SchwarzSNL022,DBLP:conf/nips/ZhuZZ00TF18}.
The feature-map-based 3D representation, known as \textit{tri-plane}, was first introduced in EG3D \cite{DBLP:conf/cvpr/ChanLCNPMGGTKKW22}. 
Since then, it has become a widely adopted 3D representation that has been extensively utilized and expanded upon by numerous 3D-related works \cite{An_2023_CVPR,Wang_2023_CVPR,Chen_2023_ICCV,DBLP:journals/corr/abs-2211-11208,Zhang_2023_ICCV,NEURIPS2022_cebbd24f,DBLP:journals/corr/abs-2309-14600}.


However, most 3D-aware face GANs utilize datasets mainly comprising frontal or near-frontal views, such as \textit{FFHQ} and \textit{CelebA}. This leads to incomplete head geometry, making these generators unsuitable to serve as 3D portrait generation priors.
%
PanoHead \cite{An_2023_CVPR}, trained on \textit{FFHQ-F}, is designed to generate full-head results using an innovative \textit{tri-grid} representation. The \textit{tri-grid} representation is derived from \textit{tri-plane}, and is designed to eliminate the ``mirrored face'' artifacts.
However, given that \textit{FFHQ-F} solely encompasses the head region, PanoHead's generation is similarly restricted, resulting in a lack of complete geometry of neck and shoulder areas.
3DPortraitGAN \cite{wu20233dportraitgan} further pushes the boundaries in this direction, facilitating 3D one-quarter headshot generation (including the head, neck, and shoulders). This is achieved by constructing a novel \textit{$\it{360}^{\circ}$PHQ} dataset, incorporating mesh-guided deformation, and implementing body pose self-learning.
%

\subsection{Diffusion-based 3D Portrait Generation}
Since the advent of the Diffusion Model \cite{ho2020denoising}, it has emerged as a crucial tool for 2D image generation. Stable Diffusion \cite{Rombach_2022_CVPR}, also known as the latent diffusion model, further expands the diffusion model to the latent space of a pre-trained autoencoder model, significantly reducing computational costs while maintaining the quality of generation.
However, training a diffusion model requires substantial amounts of training data, posing challenges for 3D content generation due to the difficulty in obtaining 3D data.
To bridge this gap between 2D and 3D space, DreamFusion \cite{DBLP:conf/iclr/PooleJBM23} introduces a novel method known as Score Distillation Sampling (SDS). SDS seeks to optimize the parameters of a differential renderer by penalizing the discrepancy between the predicted noise and the noise added to the rendered image, aiming to make the rendered image align with the image distribution learned by the diffusion model.
ProlificDreamer \cite{wang2023prolificdreamer} introduces Variational Score Distillation (VSD), which treats the 3D scene given a textual prompt as a random variable instead of a single point, effectively alleviating issues of over-saturation, over-smoothing, and low diversity encountered in SDS.

By adapting general diffusion-based text-to-3D techniques to portrait-specific designs, text-to-3D portrait generation has become a popular research topic. 
Some methods employ meshes and textures to represent 3D portraits \cite{DBLP:journals/corr/abs-2308-10899,DBLP:journals/corr/abs-2308-03610,xu2023seeavatar}, and these generation methods can be accomplished by first deforming parametric body shape models and then generating corresponding textures. 
However, these methods face challenges in generating realistic results.
Neural implicit representation \cite{DBLP:conf/eccv/MildenhallSTBRN20,DBLP:conf/nips/WangLLTKW21,10.1145/3592433}, on the other hand, can represent detailed shapes and produce realistic results, making it a popular 3D portrait representation. 
Since the inception of 3D Gaussian Splatting \cite{DBLP:journals/tog/KerblKLD23}, it has emerged as a popular representation of 3D portraits, capable of achieving real-time rendering. This method has been employed in portrait reconstruction studies \cite{wang2024gaussianhead,DBLP:journals/corr/abs-2402-03723,DBLP:journals/corr/abs-2312-04558,DBLP:journals/corr/abs-2312-03029}, as well as in text-to-3D-portrait generation \cite{DBLP:journals/corr/abs-2402-06149}.
%
Because human faces and bodies exhibit strong regularity, prior information can be extracted from a variety of human data sources to guide diffusion-based 3D portrait generation.
A stream of research employs body shape models, such as SMPL \cite{DBLP:journals/corr/abs-2308-09712,Jiang_2023_ICCV}, FLAME \cite{DBLP:journals/corr/abs-2306-03038}, imGHUM \cite{DBLP:journals/corr/abs-2306-09329}, as the initial point for text-to-3D-portrait generation. 
3D avatars can also be generated from reference images \cite{huang2023tech,10.1145/3610548.3618153,ho2023sith} by utilizing the diffusion model to generate unseen regions with seamless integration, which can be guided by a human body shape prediction network \cite{10.1145/3610548.3618153} or a back appearance hallucinator and normal predictor \cite{ho2023sith}.
Research efforts have also been directed towards editing 3D objects, including 3D portraits, utilizing the diffusion model \cite{DBLP:conf/iccv/HaqueTEHK23}.
%

%
Nonetheless, the necessity for a robust joint geometry-appearance prior persists. Exclusively considering geometry information leads to issues such as inconsistent or unrealistic texture. Additionally, implementing SDS on a coarse initialization can engender over-saturation and over-smoothing.
In this paper, we propose using 3D-aware GANs as a robust prior, which learns the joint distribution of portrait geometry and texture, thereby enhancing the process of 3D portrait generation.

\section{Methodology}
\label{sec: Methodology}

\subsection{Overview}
In this section, we present an overview of \ourname.
We first introduce our novel 3D representation named \textit{pyramid tri-grid} in Sec. \ref{sec: pyramid tri-grid}. This representation incorporates multiple-resolution features and aids in alleviating the ``grid-like'' artifact that occurs during score distillation sampling.
In Sec. \ref{sec: Full-Head 3D-aware Generator}, we describe our 3D-aware generator, 3DPortraitGAN\resizebox{0.30cm}{!}{~\protect}\xspace. This generator incorporates a 3D-aware branch into the original 2D StyleGAN backbone, allowing 3DPortraitGAN\resizebox{0.30cm}{!}{~\protect}\xspace to output the \textit{pyramid tri-grid} and furnish prior information for the subsequent process.
Next, in Sec. \ref{sec: Diffusion-based Generation}, we propose a novel text-guided 3D portrait generation algorithm. Leveraging the prior information provided by 3DPortraitGAN\resizebox{0.30cm}{!}{~\protect}\xspace, this algorithm produces 3D portraits via score distillation sampling.
Lastly, in Sec. \ref{sec: Diffusion-based Optimization}, we elaborate on the final procedure to further enhance the quality of 3D portraits. We adopt a diffusion model to refine the rendered images and utilize them to optimize the 3D portrait.

\begin{figure}[t]
  \centering
  \includegraphics[width=0.9\linewidth]{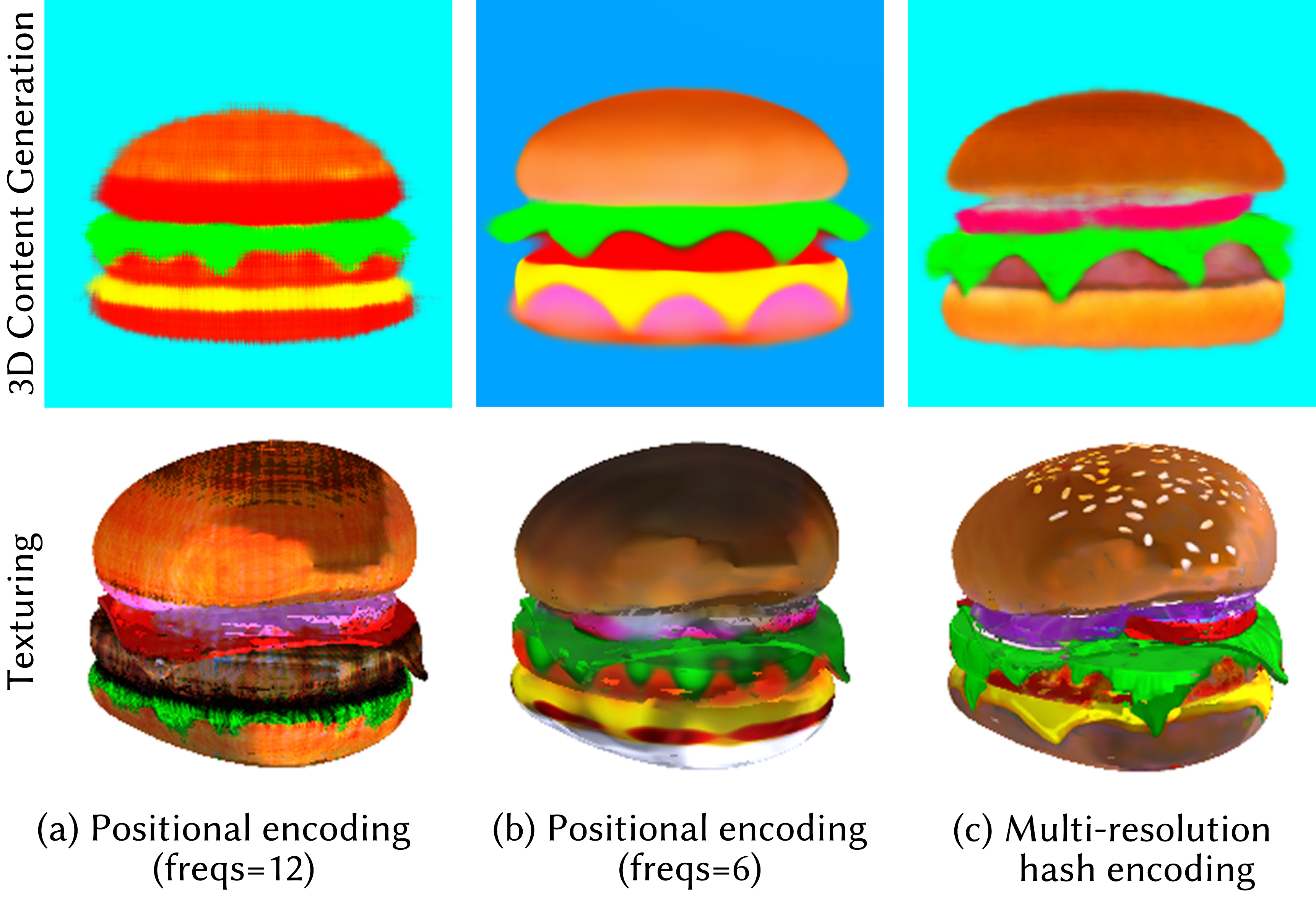}
  \caption{
  The results of score distillation sampling for 3D content generation (top) and texture generation (bottom), using positional encoding with a different number of frequencies (a,b), and multi-resolution hash encoding (c). The same prompt, ``a hamburger'', was used for a fair comparison.
  }
  \label{fig: toy}
  \vspace{-10pt}
\end{figure}

\subsection{Pyramid Tri-grid}
\label{sec: pyramid tri-grid}

3DPortraitGAN \cite{wu20233dportraitgan} is a 3D-aware generator capable of generating 3D full-head avatars, and can function as a prior for 3D portrait generation.
3DPortraitGAN draws inspiration from Panohead \cite{An_2023_CVPR} and utilizes a feature-map-based 3D representation named \textit{tri-grid} (we denote it as $T \in \mathbb{R}^{3 \times 3 \times 32 \times 256 \times 256}$) to store the features of color and density of a 3D portrait. 


During the neural rendering process, the \textit{tri-grid} is processed through a neural renderer to produce RGB images:
\begin{equation}
\begin{split}
    \label{eqn: neural renderer}
     x_{rgb} = R(T,c,w),
    \end{split}
\end{equation}
where $R$ is the neural renderer, $c$ denotes the camera parameters, and $w$ denotes the latent code. 
The \textit{tri-grid} is first rendered to a feature image $x_{feature}$ via volume rendering. Specifically, a point $\bold{x}$ is sampled on the ray and then projected onto the \textit{tri-grid} to query features, which are aggregated through summation and then input into a decoder to predict the color features and density required for volume rendering. 
Next, a ToRGB module, modulated by the latent code $w$, converts $x_{feature}$ into the RGB image $x_{rgb}$.



A straightforward 3D portrait generation approach might involve directly applying score distillation sampling to the \textit{tri-grid} generated by the 3DPortraitGAN, which already stores the information of a coarse 3D portrait. However, this results in significant ``grid-like'' artifacts (detailed in Sec. \ref{sec: pyramid_trigrid_ablation}). We attribute this phenomenon to the exclusive high-frequency information encoded in the single-resolution \textit{tri-grid} representation. 
%


We conducted a simple example to demonstrate our intuition, as shown in Fig. \ref{fig: toy}. Score distillation sampling was used in this experiment for two tasks: (1) 3D content generation\footnote{We used the stable-diffusion-2-1-base model and adopted the SDS implementation provided by \href{https://github.com/ashawkey/stable-dreamfusion}{https://github.com/ashawkey/stable-dreamfusion}.} and (2) generating texture for a given mesh\footnote{We utilized the stable-diffusion-2-1-base model and used the ``Texture phase'' of Fantasia3D from \href{https://github.com/threestudio-project/threestudio}{https://github.com/threestudio-project/threestudio}.}, with the prompt ``a hamburger.''.
To facilitate the representation of high-frequency content, we used the positional encoding strategy \cite{DBLP:conf/eccv/MildenhallSTBRN20}. As shown in Fig. \ref{fig: toy} (a), positional encoding causes ``grid-like'' artifacts. 
We then investigated two methods to mitigate the effect of high-frequency information: reducing the maximum frequency of positional encoding from 12 to 6 (Fig. \ref{fig: toy} (b)) and employing multi-resolution hash encoding \cite{DBLP:journals/tog/MullerESK22} (Fig. \ref{fig: toy} (c)).
Both methods successfully reduce ``grid-like'' artifacts.
Owing to its multi-resolution structure, hash encoding could achieve a better balance between generating highly detailed results and mitigating high-frequency noise.


Inspired by multi-resolution hash encoding, we introduce a novel 3D representation named \textit{pyramid tri-grid}, which incorporates a similar multi-resolution pyramid structure by encompassing \textit{tri-grids} of different resolutions: $\{8,16,32,64,128,256,512\}$. During feature querying, features on \textit{tri-grids} of different resolutions are aggregated through summation. We find that a channel dimension of 12 offers sufficient expressive power, thereby our \textit{pyramid tri-grid} is defined as $T^{pyr} = [T^{8} \in \mathbb{R}^{3 \times 3 \times 12 \times 8 \times 8},\cdots, T^{512} \in \mathbb{R}^{3 \times 3 \times 12 \times 512 \times 512}]$.

\begin{figure}[t]
  \centering
  \includegraphics[width=\linewidth]{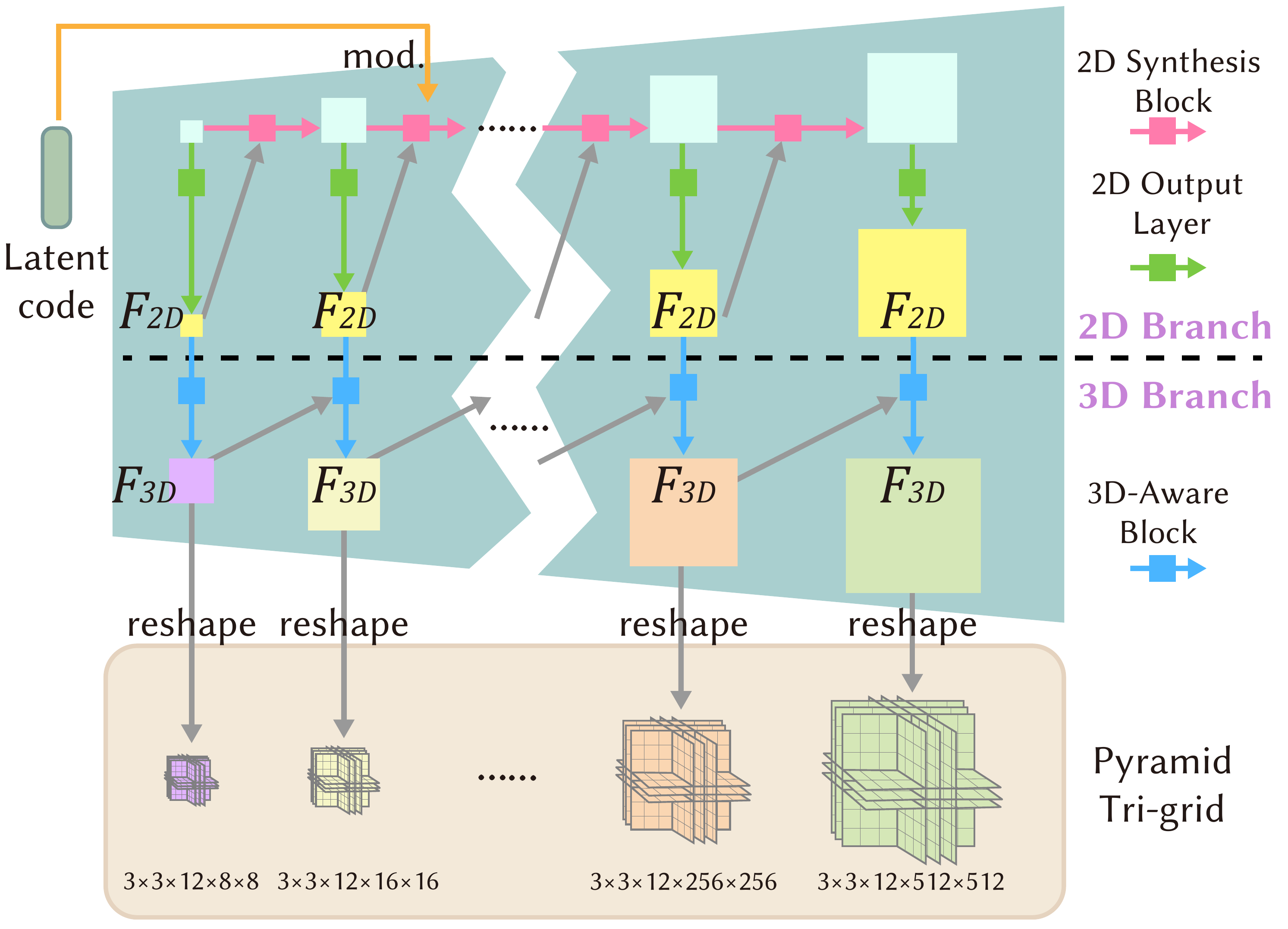}
  \caption{
   The architecture of the 3D-aware \textit{pyramid tri-grid} generator in 3DPortraitGAN\resizebox{0.30cm}{!}{~\protect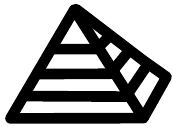}\xspace.
    The \textit{pyramid tri-grid} is composed of \textit{tri-grids} generated at different layers. For the sake of simplicity and clarity, we omit the latent code modulation applied to each block.
  }
  \label{fig: generator}
\end{figure}

\subsection{3DPortraitGAN\resizebox{0.30cm}{!}{~\protect}\xspace}
\label{sec: Full-Head 3D-aware Generator}

We subsequently incorporate the \textit{pyramid tri-grid} into 3DPortraitGAN, thereby establishing it as the foundational 3D representation. We refer to this newly developed 3D portrait generator as 3DPortraitGAN\resizebox{0.30cm}{!}{~\protect}\xspace (the network architecture is detailed in the supplementary file). 

To implement the generation of multiple-resolution \textit{pyramid tri-grid}, we propose an innovative 3D-aware \textit{pyramid tri-grid} generator, as shown in Fig. \ref{fig: generator}.
Inspired by Mimic3D \cite{Chen_2023_ICCV}, we incorporate a 3D-aware branch into the original 2D StyleGAN backbone to bolster feature communications between 3D-associated positions across different feature maps.
At each layer, the 2D branch outputs the feature map $F_{2D}$ through a 2D output layer. $F_{2D}$ is then fed into the 3D branch. 
Within the 3D branch, a 3D-aware block equipped with a $\times 2$ upsample operation processes $F_{2D}$, leading to a feature map $F_{3D}$. $F_{3D}$ is then reshaped into a \textit{tri-grid}. Additionally, $F_{3D}$ is fed to the 3D-aware block of the succeeding 3D layer.

\subsection{Diffusion-based 3D Portrait Generation}
\label{sec: Diffusion-based Generation}

We rely on the prior information provided by our 3DPortraitGAN\resizebox{0.30cm}{!}{~\protect}\xspace to initialize the text-to-3D-portrait generation process.

\begin{figure*}[t]
  \centering
  \includegraphics[width=0.99\linewidth]{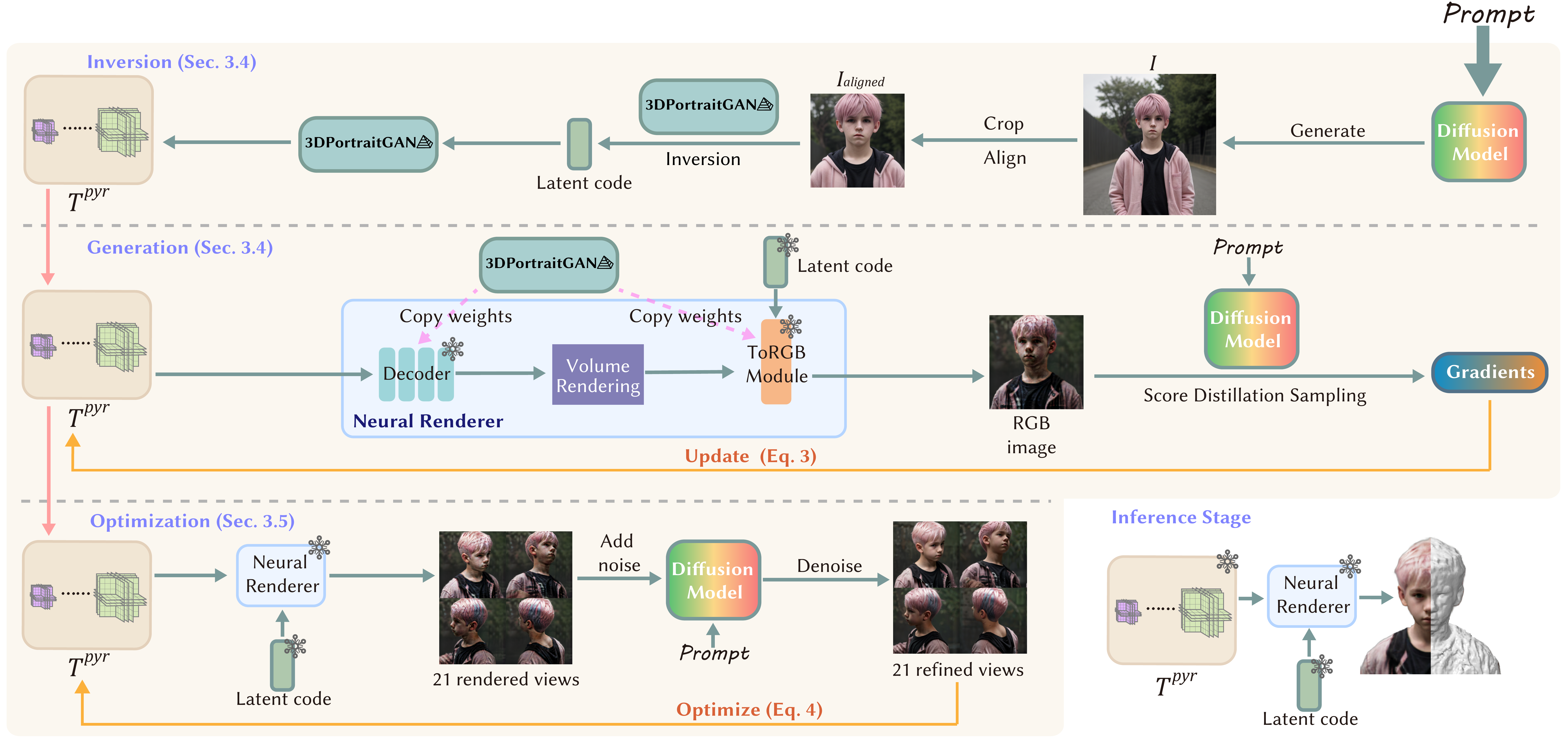}
  
  \caption{
    The 3D portrait generation pipeline of \ourname. The ``\resizebox{0.25cm}{!}{~\protect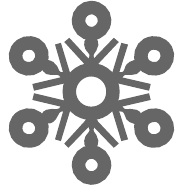}'' denotes that the submodule or representation is frozen.
  }
  \label{fig: optimization}
\end{figure*}

Specifically, as depicted in Fig. \ref{fig: optimization}, given a prompt describing a particular portrait, we first generate a random portrait image $I$ using the diffusion model, and then process $I$ to produce an aligned image $I_{aligned}$ following the alignment setting of 3DPortraitGAN\resizebox{0.30cm}{!}{~\protect}\xspace.
Subsequently, we obtain the inversion latent code $w^*$ of the aligned image through latent code optimization (detailed in the supplementary file).
Then $w^*$ is fed to the 3D-aware \textit{pyramid tri-grid} generator to obtain a \textit{pyramid tri-grid}:
\begin{equation}
\begin{split}
    \label{eqn: inversion}
    T^{pyr} &=   \mathcal{G}(w^*),
    \end{split}
\end{equation}
where $\mathcal{G}$ denotes our 3D-aware \textit{pyramid tri-grid} generator.


After that, we freeze the parameters of the neural renderer (as shown in Fig. \ref{fig: optimization}), and employ score distillation sampling to distill the knowledge of the diffusion model into the \textit{pyramid tri-grid}:
\begin{equation}
\begin{split}
       & \nabla_{\theta}\mathcal{L}_{\mathrm{SDS}}\left(\phi, x = R(T^{pyr},c,w^*)\right)\\ 
       & \triangleq   \mathbb{E}_{t, \epsilon}\left[\omega(t)\left(\hat{\epsilon}_\phi\left(z_t ; y, t\right)-\epsilon\right) \frac{\partial z_0}{\partial x} \frac{\partial x}{\partial \theta}\right].
\end{split}
\end{equation}
Here, $\phi$ represents the parameters of the U-Net network $\hat{\epsilon}_\phi$ within the latent diffusion model, while $\theta$ denotes the parameters of the \textit{pyramid tri-grid} $T^{pyr}$.
$\epsilon$ represents the random noise added to the latent $z_0$. The latent $z_0$ is derived from the rendered RGB image $x = R(T^{pyr},c,w^*)$ by feeding it into the encoder of the latent diffusion model's autoencoder. The term $\frac{\partial z_0}{\partial x}$ refers to the Jacobian of the autoencoder's encoder, playing the role in backpropagating gradients from the latent space to the RGB space. Meanwhile, $z_t$ represents the noisy latent at time $t$, obtained by adding noise to $z_0$.
$y$ denotes the text embeddings.
$\omega(t)$ serves as a weighting function that absorbs the constant $\alpha_t \textbf{I} = \partial z_t / \partial z_0$. 
The gradient $\nabla_{\theta}\mathcal{L}_{\mathrm{SDS}}\left(\phi, x = R(T^{pyr},c,w^*)\right)$ is utilized to update $\theta$ .

%

\subsection{Diffusion-based 3D Portrait Optimization}
\label{sec: Diffusion-based Optimization}

After applying score distillation sampling, the 3D portraits generated from $T^{pyr}$ still exhibit unnatural artifacts (refer to Sec. \ref{sec: Postprocess Optimization}). To address this, we propose a method to further enhance the quality of the obtained 3D portrait.

As shown in Fig. \ref{fig: optimization}, we render 21 views using $T^{pyr}$. Specifically, we sample 7 uniformly distributed azimuth angles from the range $[0^{\circ},360^{\circ}]$. For each azimuth angle, we sample 3 elevation angles from the ranges $[55^{\circ},65^{\circ}]$,  $[85^{\circ},95^{\circ}]$,  and $[115^{\circ},125^{\circ}]$,  respectively. 
These angles are then used to render 21 views by feeding $T^{pyr}$ to the neural renderer.
%
Subsequently, we add randomly sampled noise with a constant noise level to the rendered views, and employ the diffusion model to denoise these images. This step aims to yield high-quality views of the 3D portraits devoid of artifacts.
%
Finally, we compute the $L_2$ loss between the refined views and the rendered views of $T^{pyr}$ to optimize $T^{pyr}$'s parameters $\theta$:
\begin{equation}
\begin{split}
    \label{eqn: refine}
    \theta^* = \mathop{\arg\min}_\theta L_{optim} = \mathop{\arg\min}_\theta L_2\left(x^c_{refined}, R(T^{pyr},c,w^*)\right),
    \end{split}
\end{equation}
where $x^c_{refined}$ denotes the refined view rendered using camera parameters $c$. 
We optimize $T^{pyr}$'s parameters by penalizing the $L_{optim}$ loss while keeping the parameters of the neural renderer unchanged, thus obtaining the final result of our framework. 
As depicted in Fig. \ref{fig: optimization}, during the inference phase, $T^{pyr}$ represents a high-quality 3D portrait and can generate view-consistent portrait images via the neural renderer.

\begin{figure*}[t]
  \centering
  \includegraphics[width=0.9\linewidth]{comparison.pdf}
  \caption{
  Qualitative comparison to SOTA text-to-3D approaches: DreamFusion \cite{DBLP:conf/iclr/PooleJBM23},  LucidDreamer \cite{DBLP:journals/corr/abs-2311-11284}, TADA \cite{DBLP:journals/corr/abs-2308-10899}, AvatarCraft \cite{Jiang_2023_ICCV}, AvatarStudio \cite{DBLP:journals/corr/abs-2311-17917}, HumanGaussian \cite{DBLP:journals/corr/abs-2311-17061}, AvatarVerse \cite{DBLP:journals/corr/abs-2308-03610}, HumanNorm \cite{DBLP:journals/corr/abs-2310-01406}, SEEAvatar \cite{xu2023seeavatar}, TECA \cite{zhang2023textguided}, and our method. The input prompt is presented at the top.
  }
  \label{fig: comparison}
\end{figure*}

\section{Results}
\label{sec: Results}

\subsection{Visual Results} 
\begin{figure*}[t]
  \centering
  \includegraphics[width=0.97\linewidth]{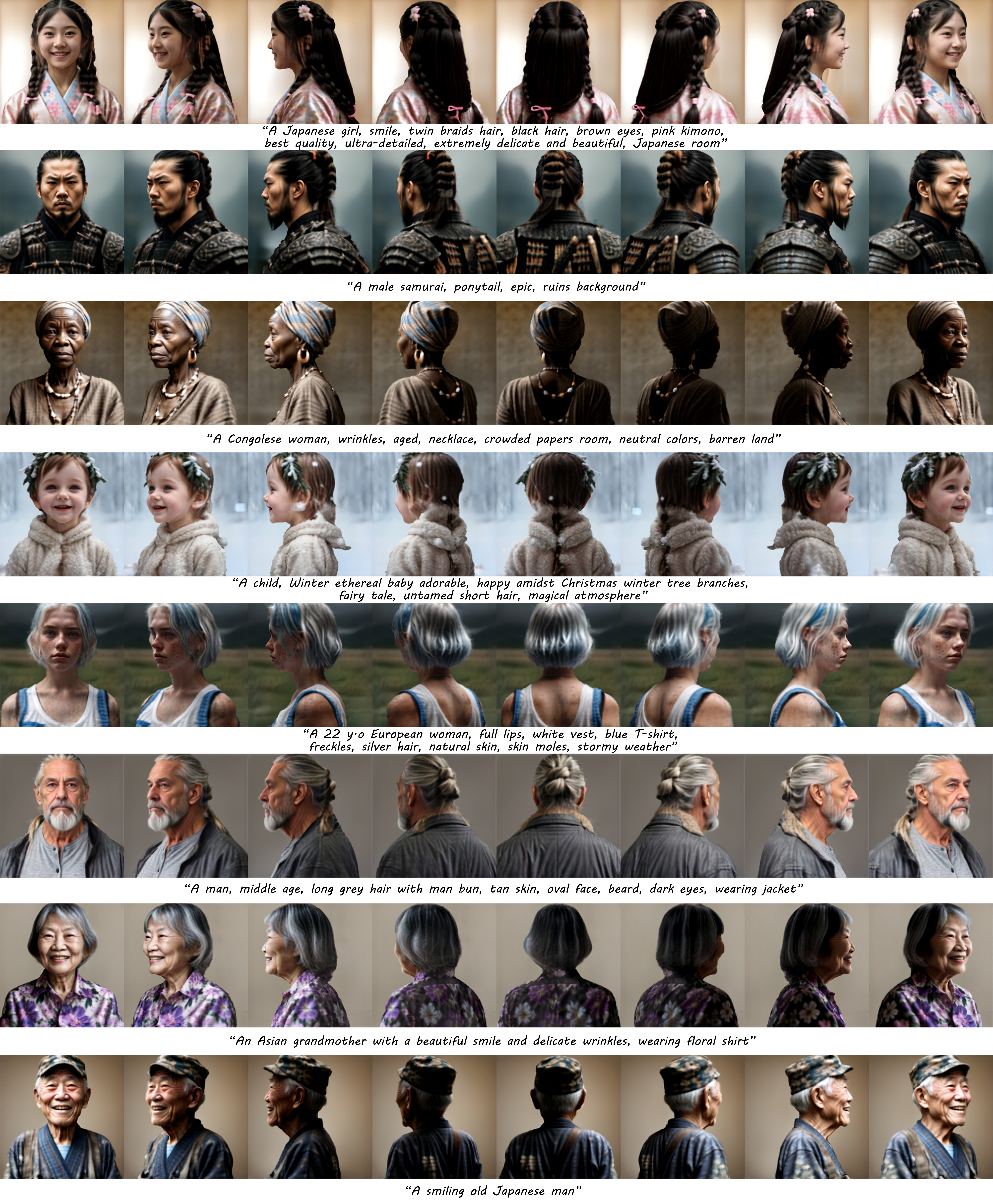}
  
  \caption{Examples of 3D portraits generated using \ourname.}
  \label{fig: results-3}
\end{figure*}

\begin{figure*}[t]
  \centering
  \includegraphics[width=0.97\linewidth]{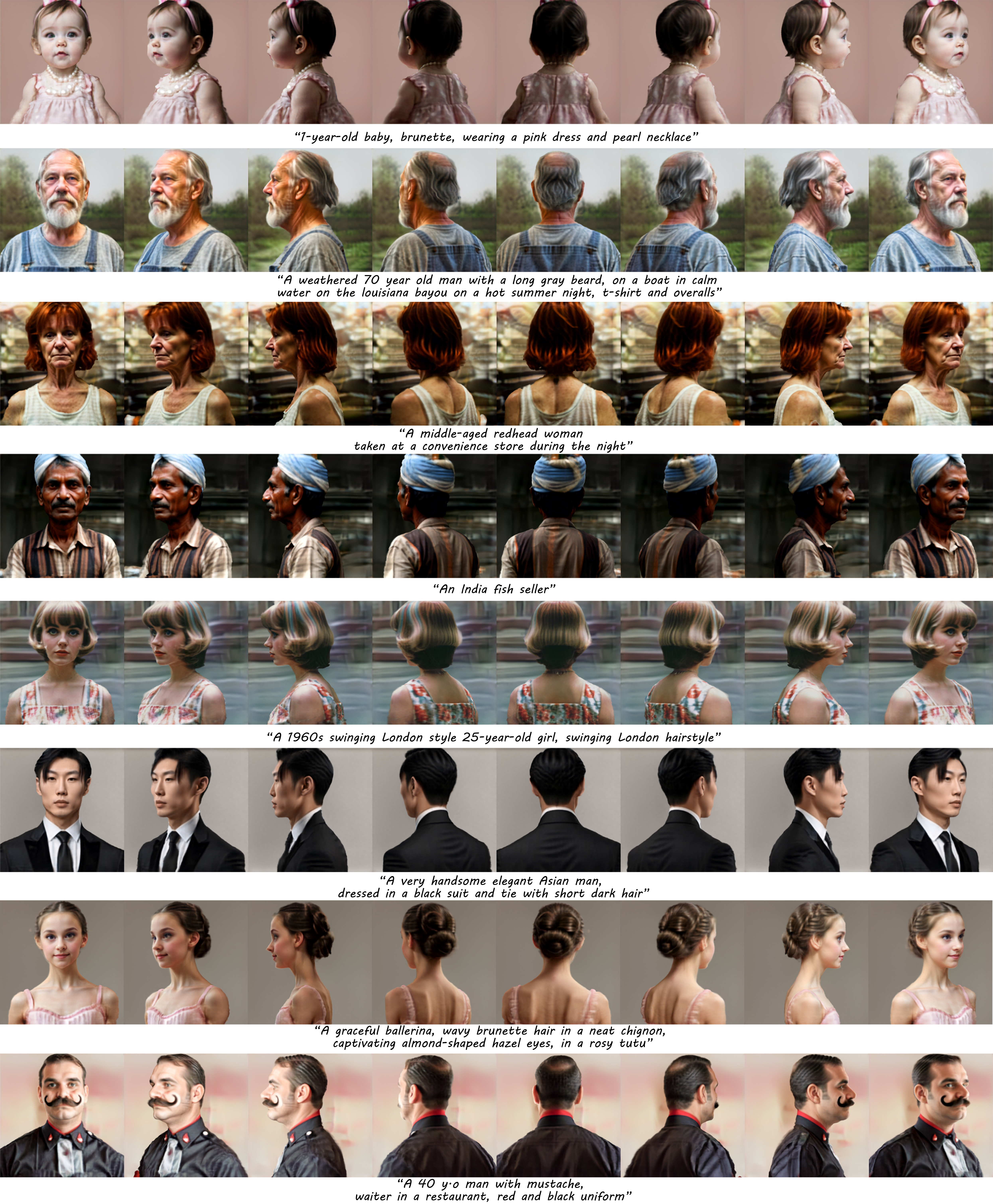}
  
  \caption{Examples of 3D portraits generated using \ourname.}
  \label{fig: results-2}
\end{figure*}

Fig. \ref{fig: results-3} and Fig. \ref{fig: results-2} present a set of text-to-3D-portrait results generated by \ourname. 
The background for each instance is generated by the background generator in 3DPortraitGAN\resizebox{0.30cm}{!}{~\protect}\xspace utilizing the inversion latent code. The results are generated in the canonical space.
Considering that the results of our 3DPortraitGAN\resizebox{0.30cm}{!}{~\protect}\xspace predominantly cover the upper body region, all our prompts commence with the phrase ``upper body photo''. Additionally, we include terms such as ``8k UHD'', ``high quality'', and ``cinematic'' in the input prompts to facilitate the generation of more stable, high-quality images. 
We omit these terms that are not directly pertinent to the semantics of 3D portraits from the figures to avoid redundancy.

The results illustrate the capability of \ourname to generate 3D portraits that are highly consistent with the input text. 
Owing to the robust prior information from 3DPortraitGAN\resizebox{0.30cm}{!}{~\protect}\xspace, the generated portraits are realistic and high-quality, successfully avoiding problems such as over-saturation and the Janus problem.
Moreover, \ourname is able to generate a diverse range of 3D portraits, encompassing various races, ages, genders, hairstyles, and so forth.
More results are presented in the supplementary file.

\subsection{Runtime} 

\ourname necessitates 0.5 hours on a single NVIDIA RTX 4090 GPU with 24 GB of memory to generate a 3D portrait. The image inversion process takes approximately 4 minutes. The 3D portrait generation process requires about 22 minutes. The 3D portrait optimization process consumes around 5 minutes.
\ourname can also operate on a GPU with only 12 GB of memory, such as the NVIDIA RTX 3080Ti, with a processing time requirement of 1.5 hours. The image inversion process takes approximately 6.5 minutes. The 3D portrait generation process requires about 50 minutes. The 3D portrait optimization process consumes around 30 minutes.

\begin{figure*}[t]
  \centering
  \includegraphics[width=0.99\linewidth]{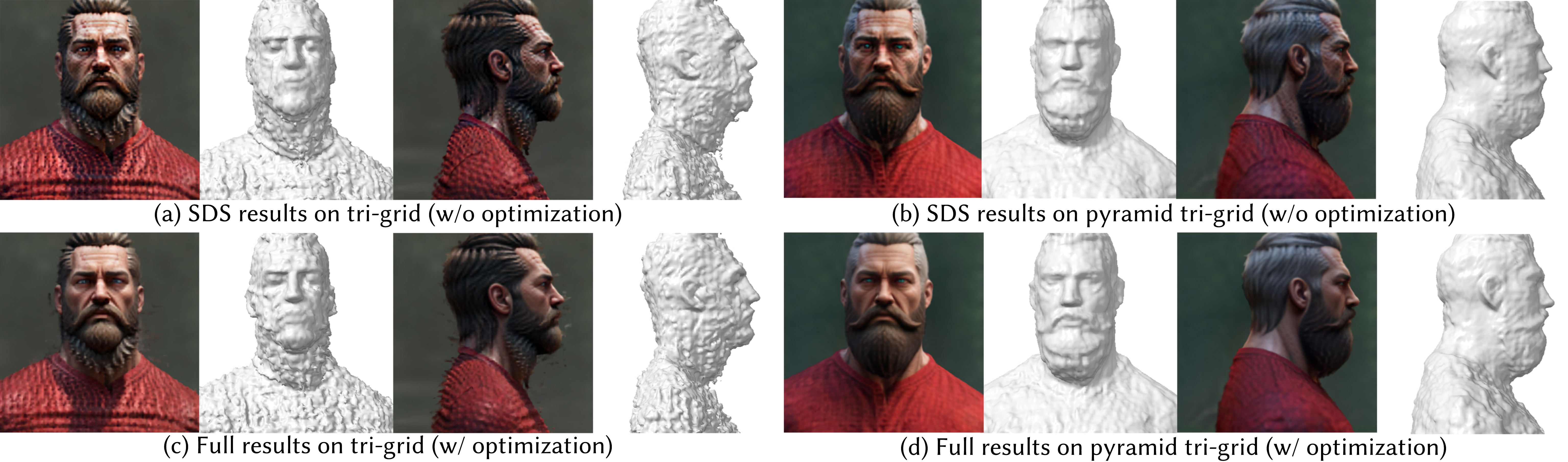}
  \caption{
  The \textit{pyramid tri-grid} is crucial for alleviating the ``grid-like'' artifacts. 
  We showcase renderings of results obtained utilizing the two 3D representations (w/ and w/o optimization), accompanied by shapes extracted using Marching Cubes.
  }
  \label{fig: pyramid_trigrid_ablation}
\end{figure*}

\section{Comparison}
\label{sec: Comparison}

We compare our \ourname with text-to-3D approaches focused on general objects, such as DreamFusion \cite{DBLP:conf/iclr/PooleJBM23} and LucidDreamer \cite{DBLP:journals/corr/abs-2311-11284}, and approaches focused on 3D portraits, such as TADA \cite{DBLP:journals/corr/abs-2308-10899}, AvatarCraft \cite{Jiang_2023_ICCV}, AvatarStudio \cite{DBLP:journals/corr/abs-2311-17917}, HumanGaussian \cite{DBLP:journals/corr/abs-2311-17061}, AvatarVerse \cite{DBLP:journals/corr/abs-2308-03610}, HumanNorm \cite{DBLP:journals/corr/abs-2310-01406}, SEEAvatar \cite{xu2023seeavatar} and TECA \cite{zhang2023textguided}.


\begin{table}[t]
\caption{Quantitative comparison results. The results with color \textcolor{moreprompt}{$\blacksquare$} are derived from 25 distinct input prompts, while those with color \textcolor{oneprompt}{$\blacksquare$} are derived from a single input prompt due to the inaccessibility of some methods.}
\scalebox{0.82}{ 
\begin{tabular}{@{}ccccccccp{0.5cm}<{\centering}@{}}
\toprule
\multirow{2}{*}{Method} & \multicolumn{2}{c}{Quality$\uparrow$} & \multicolumn{2}{c}{Alignment$\uparrow$} & \multicolumn{2}{c}{\multirow{2}{*}{FID$\downarrow$}} & \multicolumn{2}{c}{\multirow{2}{*}{CLIP Score$\uparrow$}} \\
                        & \multicolumn{2}{c}{(User Study)}      & \multicolumn{2}{c}{(User Study)}        & \multicolumn{2}{c}{}                                 & \multicolumn{2}{c}{}                                      \\ \midrule
DreamFusion      &\makebox[0.02\textwidth]{1.10}\cellcolor{moreprompt}   & \makebox[0.02\textwidth]{1.35}\cellcolor{oneprompt}  & \makebox[0.02\textwidth]{1.54}\cellcolor{moreprompt} & \makebox[0.02\textwidth]{2.95}\cellcolor{oneprompt}  & \makebox[0.03\textwidth]{285.5}\cellcolor{moreprompt} & \makebox[0.03\textwidth]{336.2}\cellcolor{oneprompt}   & \makebox[0.02\textwidth]{0.61}\cellcolor{moreprompt} & \makebox[0.02\textwidth]{\textbf{0.76}}\cellcolor{oneprompt}        \\
LucidDreamer     & \cellcolor{moreprompt}2.28  & \cellcolor{oneprompt}1.75  & \cellcolor{moreprompt}3.36   & \cellcolor{oneprompt}2.85 &\cellcolor{moreprompt}202.5 &\cellcolor{oneprompt}182.6 & \cellcolor{moreprompt}0.65 &\cellcolor{oneprompt}0.67    \\
TADA             & \cellcolor{moreprompt}2.57   & \cellcolor{oneprompt}1.35  & \cellcolor{moreprompt}3.24  & \cellcolor{oneprompt}2.45 &\cellcolor{moreprompt}197.2 &\cellcolor{oneprompt}180.5 & \cellcolor{moreprompt}0.68 &\cellcolor{oneprompt}0.68     \\
AvatarCraft      & \cellcolor{moreprompt}1.25  & \cellcolor{oneprompt}1.05   & \cellcolor{moreprompt}1.40   & \cellcolor{oneprompt}1.75 &\cellcolor{moreprompt}248.9 &\cellcolor{oneprompt}341.8  & \cellcolor{moreprompt}0.57 &\cellcolor{oneprompt}0.51      \\
HumanGaussian    & \cellcolor{moreprompt}3.30   & \cellcolor{oneprompt}3.40   & \cellcolor{moreprompt}\underline{3.66}   & \cellcolor{oneprompt}\underline{3.85} &\cellcolor{moreprompt}203.9 &\cellcolor{oneprompt}214.4 & \cellcolor{moreprompt}\underline{0.73} &\cellcolor{oneprompt}0.66 \\
HumanNorm        & \cellcolor{moreprompt}\underline{3.41}   & \cellcolor{oneprompt}\underline{3.70}  & \cellcolor{moreprompt}2.90   & \cellcolor{oneprompt}3.70 &\cellcolor{moreprompt}\underline{163.1} &\cellcolor{oneprompt}\underline{161.8} & \cellcolor{moreprompt}0.67 &\cellcolor{oneprompt}0.71   \\
AvatarStudio     & \cellcolor{moreprompt}N/A    & \cellcolor{oneprompt}3.10   & \cellcolor{moreprompt}N/A   & \cellcolor{oneprompt}3.70 &\cellcolor{moreprompt}N/A &\cellcolor{oneprompt}204.6 & \cellcolor{moreprompt}N/A &\cellcolor{oneprompt}0.71       \\
AvatarVerse      & \cellcolor{moreprompt}N/A    & \cellcolor{oneprompt}1.80  & \cellcolor{moreprompt}N/A   & \cellcolor{oneprompt}2.40 &\cellcolor{moreprompt}N/A &\cellcolor{oneprompt}211.4 & \cellcolor{moreprompt}N/A&\cellcolor{oneprompt}0.71  \\
SEEAvatar        & \cellcolor{moreprompt}N/A   & \cellcolor{oneprompt}3.45    & \cellcolor{moreprompt}N/A   & \cellcolor{oneprompt}4.15 &\cellcolor{moreprompt}N/A &\cellcolor{oneprompt}195.0 & \cellcolor{moreprompt}N/A&\cellcolor{oneprompt}0.74    \\
TECA             & \cellcolor{moreprompt}N/A     & \cellcolor{oneprompt}2.40   & \cellcolor{moreprompt}N/A     & \cellcolor{oneprompt}1.95 &\cellcolor{moreprompt}N/A &\cellcolor{oneprompt}169.9  & \cellcolor{moreprompt}N/A&\cellcolor{oneprompt}0.71    \\
Ours             & \cellcolor{moreprompt}\textbf{4.77}   & \cellcolor{oneprompt}\textbf{4.75}   & \cellcolor{moreprompt}\textbf{4.69} & \cellcolor{oneprompt}\textbf{4.90} &\cellcolor{moreprompt}\textbf{110.6} &\cellcolor{oneprompt}\textbf{148.5}  & \cellcolor{moreprompt}\textbf{0.80} &\cellcolor{oneprompt}\underline{0.74}     \\
\bottomrule
\end{tabular}

}
\label{tab: user_study-1}
\end{table}

\subsection{Qualitative Comparison}
As shown in Fig. \ref{fig: comparison},  we present a qualitative comparison between the SOTA methods and ours. 
To concentrate on the upper body region, we cropped some full-body results.

%

The outputs from DreamFusion and AvatarCraft display the Janus problem.
LucidDreamer generates distorted geometry.
TADA generates distorted and over-saturated results, lacking in realism.
Both AvatarStudio and HumanGaussian fall short in terms of detail, suffering from unrealistic geometry and appearance.
AvatarVerse produces unrealistic geometry and appearances.
TECA does not generate the black dress in the input prompt and produces an unnatural hairstyle.
In contrast, our \ourname demonstrates superior performance in terms of result quality, generating high-quality and realistic 3D portraits. 
Furthermore, when compared to methods that represent 3D portraits using meshes and textures (such as TADA, AvatarVerse, HumanNorm, and SEEAvatar), our neural rendering-based framework produces more realistic results and exhibits the capability to represent complex shapes and materials.

\subsection{Quantitative Comparison}
We conducted a user study to quantitatively compare our \ourname to SOTA alternatives. 
We asked 20 participants to evaluate videos rendered from the 3D portraits generated by different methods and assign scores (ranging from 1 to 5) to those videos based on two criteria: overall quality and alignment with the input prompt. 
Moreover, the quality of the rendered views is further evaluated using the Fr\'echet Inception Distance (FID) \cite{DBLP:conf/nips/HeuselRUNH17} by comparing the distribution's distance between images generated by the diffusion model and rendered views. Additionally, we use the CLIP Score \cite{DBLP:conf/emnlp/HesselHFBC21} to quantify the semantic alignment between the prompts and the rendered views. For more details on quantitative experiments, please refer to the supplementary file.


Tab. \ref{tab: user_study-1} presents the average scores of DreamFusion, LucidDreamer, TADA, AvatarCraft, HumanGaussian, HumanNorm and ours across 25 different prompts. 
As the codes for AvatarStudio, AvatarVerse, SEEAvatar, and TECA are not publicly accessible, we also compare these methods with ours and other open-source ones using a single prompt (the one in Fig. \ref{fig: comparison}).

The results of user studies, FID, and CLIP Score demonstrate that \ourname outperforms both general object generation and 3D avatar generation methods, achieving the highest overall result quality and the most accurate alignment with the prompt's semantics.
Notably, DreamFusion achieves a greater CLIP Score when assessed with a single prompt, largely due to its Janus problem in that case (as illustrated in Fig. \ref{fig: comparison}). This issue leads most of its rendered view to better align with the ``face'' term in the input prompt.

\section{Ablation Study}
\label{sec: Ablation Study}

\subsection{Pyramid Tri-grid}
\label{sec: pyramid_trigrid_ablation}



In Sec. \ref{sec: pyramid tri-grid}, we discussed the limitation of employing \textit{tri-grid}, as it generates a noticeable ``grid-like'' artifact in the outputs. This effect is depicted in Fig. \ref{fig: pyramid_trigrid_ablation}, using the same input prompt across all instances.

Following the generation method in Sec. \ref{sec: Diffusion-based Generation}, we use 3DPortraitGAN (which utilizes the \textit{tri-grid} representation) to provide prior, as displayed in Fig. \ref{fig: pyramid_trigrid_ablation} (a). 
Subsequently, to illustrate the impact of using \textit{tri-grid} in the final results, we apply the optimization process outlined in Sec. \ref{sec: Diffusion-based Optimization} to the SDS results on \textit{tri-grid}. This is demonstrated in Fig. \ref{fig: pyramid_trigrid_ablation} (c).
To ensure a fair comparison, we trained a new instance of 3DPortraitGAN\resizebox{0.30cm}{!}{~\protect}\xspace that generates a \textit{pyramid tri-grid} with resolutions $\{8,16,32,64,128,256\}$, ensuring alignment with the \textit{tri-grid} in 3DPortraitGAN (with a resolution of $256$). 
Fig. \ref{fig: pyramid_trigrid_ablation} (b) illustrates the SDS results obtained using \textit{pyramid tri-grid}. 
The optimization process is also applied to display the final results with our \textit{pyramid tri-grid} in Fig. \ref{fig: pyramid_trigrid_ablation} (d).

The comparison reveals that using the \textit{tri-grid} results in a grid-like texture, particularly noticeable on the T-shirt. This artifact is even preserved in the final results after employing the optimization process. In contrast, our \textit{pyramid tri-grid} delivers a smoother, more realistic rendering with significantly reduced noise.

\subsection{GANs Prior and Optimization}
\label{sec: Postprocess Optimization}

\begin{figure}[t]
  \centering
  \includegraphics[width=0.99\linewidth]{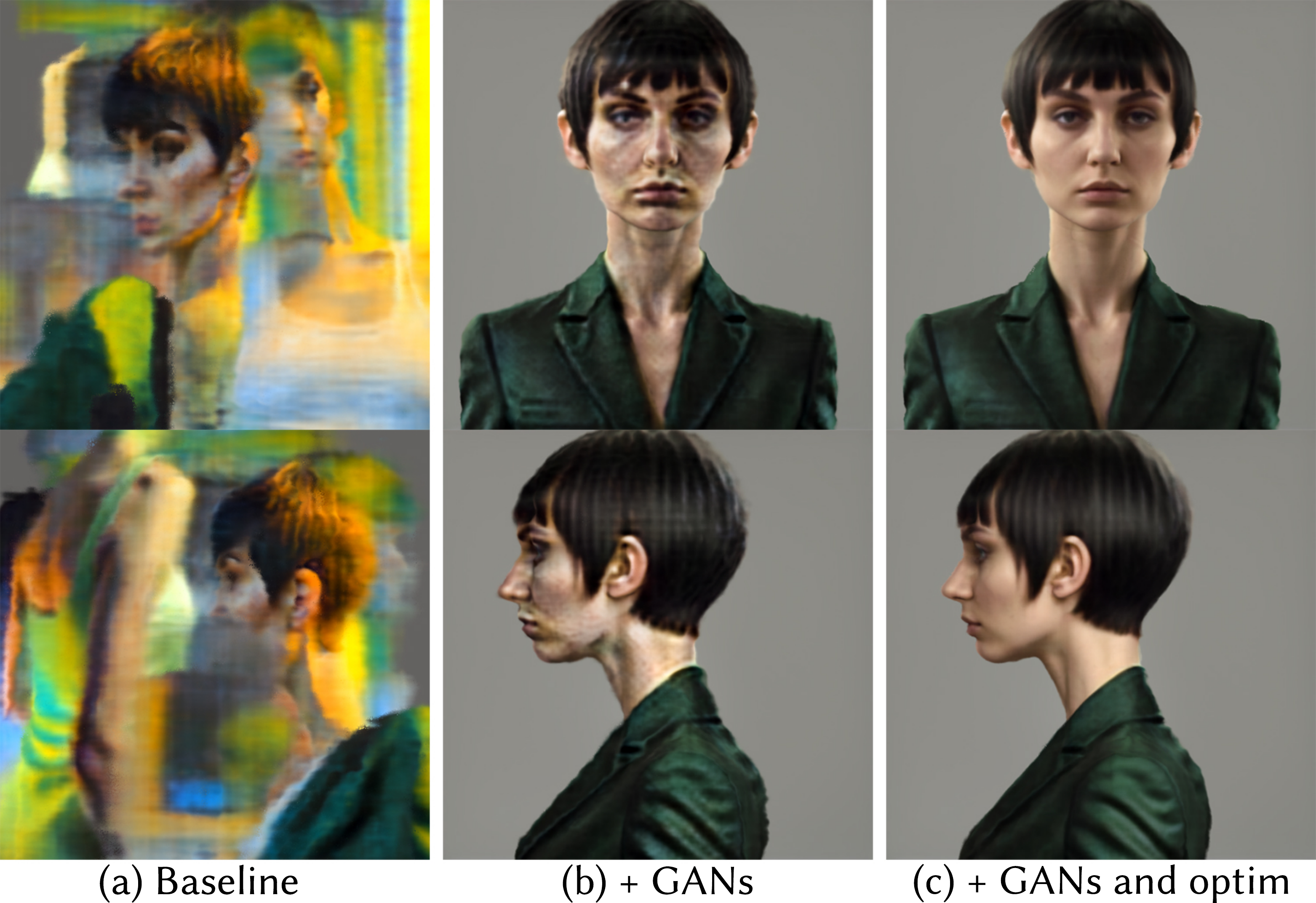}
  \caption{
  Comparison of 3D portraits generated (a) without GANs prior and optimization (baseline), (b) with GANs prior but without optimization, and (c) with both GANs prior and optimization (our complete method).
  }
  \label{fig: optimization_ablation} 
\end{figure}

The effectiveness of our GANs prior and optimization strategy is demonstrated in Fig. \ref{fig: optimization_ablation}. The Janus problem occurs when SDS is applied directly to a randomly initialized \textit{pyramid tri-grid} without GANs prior and optimization (Fig. \ref{fig: optimization_ablation} (a)). Applying the GANs prior without optimization (Fig. \ref{fig: optimization_ablation} (b)) results in an unnatural color in the outputs. The results of our full method are generated using both GANs prior and optimization (Fig. \ref{fig: optimization_ablation} (c)), displaying improved realism in geometry and appearance.


\section{Discussion}
\label{sec: Discussion}

Despite its notable performance, our method does have limitations.

\begin{figure}[t]
  \centering
  \includegraphics[width=0.99\linewidth]{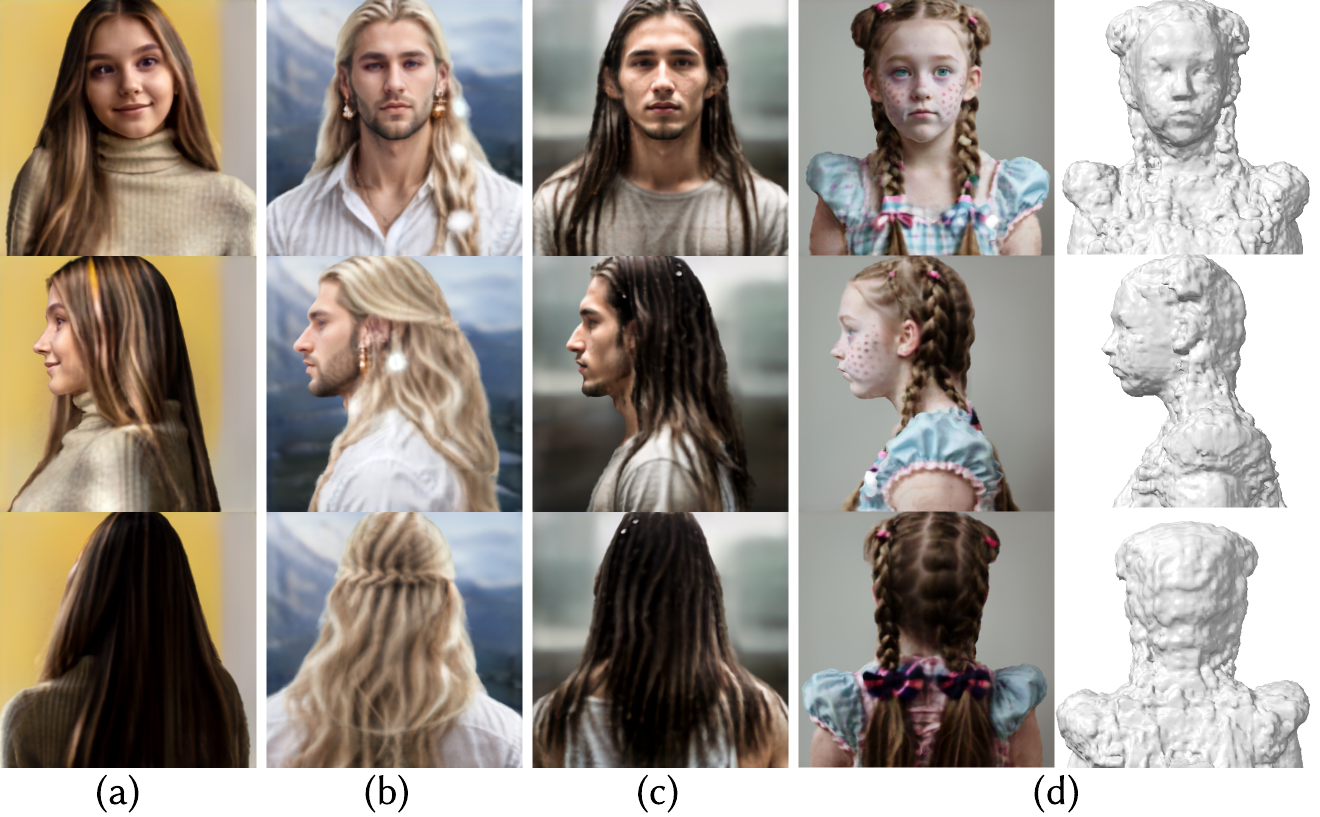}
  \caption{Some failure cases of \ourname: (a) Distortions, (b) unexpected snow, (c) semantic inconsistency, and (d) incorrect geometry.}
  \label{fig: limitation}

\end{figure}

Firstly, some 3D portraits generated by our 3DPortraitGAN\resizebox{0.30cm}{!}{~\protect}\xspace aren't entirely canonical due to the inversion method. These distortions won't be rectified during text-to-3D-portraits generation and are evident in the final results (see samples in Fig. \ref{fig: limitation} (a)).
This issue can be addressed by imposing additional constraints.

Secondly, certain semantic attributes of the background in the input prompt can influence the final result. As shown in the samples in Fig. \ref{fig: limitation} (b), the inclusion of ``snow mountain background'' in the input prompt results in unexpected snow appearing on the hair.

Thirdly, some of our results still grapple with semantic inconsistency. For instance, as seen in the samples in Fig. \ref{fig: limitation} (c), the man appears to be wearing a T-shirt from a frontal view, but seems to be wearing a vest from a rear view. This issue could potentially be addressed by employing a back-appearance hallucinator. 

Fourthly, \ourname may generate inaccurate geometry, as illustrated in Fig. \ref{fig: limitation} (d). The girl's braid is hardcoded as texture, yet it isn't accurately represented in the geometry. This issue could potentially be addressed by incorporating a more robust geometry-aware diffusion model, such as HyperHuman \cite{DBLP:journals/corr/abs-2310-08579}.

Moreover, \ourname is constrained to generating results that only include the head, neck, and shoulders. This limitation is imposed by the training dataset and the structure of 3DPortraitGAN\resizebox{0.30cm}{!}{~\protect}\xspace. This constraint could potentially be overcome by employing a 3D full-body generator in the future.

Finally, some of our results still exhibit grid artifacts. We empirically identified a trade-off between the richness of detail in our results and the reduction of grid-like artifacts. We discovered that a higher learning rate for the high-resolution \textit{tri-grid} within our \textit{pyramid tri-grid} exacerbated the grid-like artifacts, while a lower learning rate resulted in blurrier results. To achieve a balance, we choose an appropriate learning rate (detailed in the supplementary material) that preserves details and subsequently use the optimization process to eliminate any remaining artifacts.

While not constituting a technical constraint, the high-quality 3D portraits produced by our \ourname could potentially engender ethical concerns. We would like to underscore the importance of considering the potential ethical implications associated with realistic 3D portrait generation. 
The utilization of our method for the generation of realistic 3D portraits of a specific individual should be contingent on obtaining consent from the respective individual.

\section{Conclusion}
\label{sec: Conclusion}
This paper presented \ourname, an innovative text-to-3D-portrait framework that utilizes the robust prior information provided by 3D-aware GANs.
We began by training a 3D portrait generator capable of producing $360^{\circ}$ canonical 3D portraits while incorporating a novel 3D representation to circumvent the interference of exclusive high-frequency information.
Then, drawing from the prior information of the 3D portrait generator, given a text prompt, we initially project a randomly generated image aligned with the prompt into our 3D portrait generator's latent space. The resulting latent code is used to synthesize a 3D representation. Subsequently, we distill the knowledge of the diffusion model into the 3D representation through score distillation sampling.
To further enhance the quality of the 3D portrait, we apply the diffusion model to process the rendered images of the 3D portrait. The improved rendered images are then used as training data to optimize the 3D representation.
We perceive our work as an exciting step forward in the domain of 3D portrait generation, and we hope it will inspire future research in areas such as talking heads and 3D avatar reconstruction.


\bibliographystyle{ACM-Reference-Format}
\bibliography{main-bibliography}

\end{document}